# Cascading Modular Network (CAM-Net) for Multimodal Image Synthesis

**Shichong Peng, Alireza Moazeni, Ke Li**
APEX Lab
School of Computing Science
Simon Fraser University
`{shichong_peng,seyed_alireza_moazenipourasil,keli}@sfu.ca`

## Abstract

Deep generative models such as GANs have driven impressive advances in conditional image synthesis in recent years. A persistent challenge has been to generate diverse versions of output images from the same input image, due to the problem of mode collapse: because only one ground truth output image is given per input image, only one mode of the conditional distribution is modelled. In this paper, we focus on this problem of multimodal conditional image synthesis and build on the recently proposed technique of Implicit Maximum Likelihood Estimation (IMLE). Prior IMLE-based methods required different architectures for different tasks, which limit their applicability, and were lacking in fine details in the generated images. We propose CAM-Net, a unified architecture that can be applied to a broad range of tasks. Additionally, it is capable of generating convincing high frequency details, achieving a reduction of the Fréchet Inception Distance (FID) by up to $45.3\%$ compared to the baseline. More results and code are available at `https://niopeng.github.io/CAM-Net/`.

## 1 Introduction

With the advent of deep learning, image synthesis has witnessed incredible progress in recent years. Advances in deep generative models such as GANs [21, 31, 36, 8, 37], VAEs [74, 11, 59, 41], autoregressive models [63, 77, 76], normalizing flows [16, 42, 40], energy-based models [1, 19, 54], score-based models [29, 67, 27, 33] have driven remarkable improvement in the fidelity of generated images with each successive generation, and it is now possible to synthesize novel and photorealistic images with these approaches.

These successes have paved the way to the consideration of more challenging problems. Conditional image synthesis problems are often encountered in practice, where the goal is to generate images given some input that specifies what kinds of images should be generated. The conditioning input can be either discrete labels, like an object category label, or continuous labels, like another image. Among the latter class of problems, unimodal image synthesis is traditionally considered, which aims to generate an output image given an input image. In this paper, we consider the more challenging problem of *multimodal* conditional image synthesis, where the goal is to generate *multiple* plausible output images for the same given input image.

As a concrete example, suppose we are given a grayscale image of an apple and would like to generate colour versions of the image. The apple could be red or green, and the input grayscale image does not determine whether the apple should be red or green. Whereas the goal of unimodal synthesis is to generate either a red or a green apple, the goal of *multimodal* synthesis is to generate both a red apple and a green apple.

An inherent challenge of multimodal synthesis is that only one ground truth output image is observed for each input image, even though multiple output images need to be generated for the input image. A method for multimodal synthesis must therefore go *beyond* just the provided ground truth image. So, in addition to considering traditional measures of success like image fidelity, we must also consider the diversity of generated images for the same input image.

Extending GAN-based approaches to the multimodal setting has proven to be challenging [31, 96] – due to mode collapse, the generator tends to generate identical samples for the same input and ignores the latent noise. A recent method [47] takes a different approach by extending an alternative generative modelling technique known as Implicit Maximum Likelihood Estimation (IMLE) [46] to the conditional synthesis setting. While it shows promise in terms of generation diversity, it exhibits several major limitations: (1) the fidelity of generated images is lacking, (2) it used a different dedicated architecture for each of the two tasks it considered, which limits its applicability to other tasks, (3) a large number of samples need to be drawn during training to attain high generation quality, which slows down training.

In this paper, we propose a new method for multimodal conditional image synthesis called Cascading Modular Network (CAM-Net) which addresses the aforementioned issues. Our contribution is three-fold:

1. We improve the fidelity of generated images significantly by leveraging cues at multiple scales and introducing intermediate supervision
2. We devise a single unified architecture that works well for a diverse range of tasks and is easily extensible to different output resolutions
3. We propose a new sampling scheme for IMLE which attains significantly greater efficiency

We demonstrate CAM-Net significantly outperforms the prior multimodal method [47] in terms of both fidelity and diversity on a variety of tasks. Moreover, we show CAM-Net achieves superior image fidelity compared to leading unimodal methods.

## 2  Related Work

Image synthesis can be categorized into unconditional setting and conditional setting. In the unconditional setting, the goal is to learn the distribution of natural images and generate new images from the distribution. Many methods have been proposed, such as autoregressive models [63, 77, 76], VAEs [41, 74, 11, 59], GANs [21, 36, 8, 37], normalizing flows [16, 42, 40], energy-based models [1, 19, 54], score-based models [67, 27, 33] and IMLE [46].

We focus on the conditional setting in this paper, where the goal is to generate new images as described by the conditioning input. Conditional image synthesis can be further divided into class conditioning [51, 55] where the output is conditioned on class labels, and image conditioning which takes an image as conditional input. The latter is often viewed as an image-to-image translation problem, which is most closely related to the setting we consider. General image-to-image translation method can be supervised or unsupervised. In the supervised setting, each image in the source domain corresponds to a image in the target domain. Many methods have been proposed; for example, Pix2Pix [31] used an $\ell_1$ loss and a VGG feature loss and PLDT [85] added a pair-wise discriminator loss to determine the association between images from different domains. In the unsupervised setting, images in the source domain may not correspond to images in the target domain. Some examples of methods in this category include CycleGAN [94] and DualGAN [84], which adopted a bi-directional reconstruction loss to both source and target domains. DTN [72] also enforced reconstruction loss on the source domain, but the loss operates on high level feature level instead of raw pixel level. Similar to cycle-consistency loss, UNIT [49] added VAE to CoGAN [50] with the assumption that source and target domains share the same latent space. Additionally, auxiliary classifier [12, 7] can be used to categorize an image to belonged domain and improves discriminator, thereby enhances unsupervised translation quality. Additionally, some methods explore multimodality such as BicycleGAN [95] which used a hybrid approach of conditional VAE-GAN and latent regressor [20, 17], and MUNIT [28] which combines UNIT with disentangled representation learning [10, 26].

In terms of tasks, we consider four specific ones: super-resolution, image colourization, image synthesis from scene layouts and image decompression. There is a large body of work on super-resolution, most of which consider upscaling factors of $2 - 4\times$. See [83, 52, 81] for comprehensive surveys. Many methods regress to the high-resolution image directly and differ widely in the architecture [18, 39, 43, 71, 22, 93, 13, 48]. Various conditional GANs have also been developed for the problem [45, 62, 87, 80, 87, 57, 82, 79, 65]. Because these methods can only output one image for each input, they cannot handle multimodality. A prior IMLE-based method [47] tried to address this issue but suffer from a lack of fine details in the generated images.

For image colourization, [4] provides a detailed overview of different methods. The most relevant approaches are plain network and multi-path network. In the former case, [91] used a simple convolutional network and turned the problem into a classification task where different output classes represents different colours. In the latter case, [30] adopted a multi-branch network to extract details of various levels, [44] used VGG features to train a model that predicts the hue and chroma distributions for each pixel.

For image synthesis from scene layouts, some methods take input of a scene graph [24, 32] while others take a more detailed semantic map as input. Many GAN-based methods have been developed for the latter case such as AL-CGAN[34], OC-GAN [70], LostGAN [69] and LostGAN-V2 [68]. However, these models are all unimodal methods. A previous IMLE-based method [47] produces diverse results, but fails to generate high fidelity images.

There is relatively little work on image decompression to our knowledge. [5] targets DCT-based compression methods and explicitly models the quantization errors of DCT coefficients. More work was done on learned image compression [3, 2], which changes the encoding of the compressed image itself. Another related area is image denoising, see [73] for a survey of deep learning methods. Many of the methods uses a ResNet [23] backbone or other variants of deep CNNs [88, 89, 90, 78].

## 3 Background

Consider a generator network $f_\theta$ parameterized by $\theta$ that takes in two inputs, the input image $\mathbf{x}$ and a latent noise vector $\mathbf{z}$ drawn from a standard Gaussian $\mathcal{N}(0, \mathbf{I})$, and produces an image $\widehat{\mathbf{y}}$ as output. To train such a network, we can use a conditional GAN (cGAN), which adds a discriminator network that tries to tell apart the observed output $\mathbf{y}$ and the generated output $\widehat{\mathbf{y}}$. The generator is trained to make its output $\widehat{\mathbf{y}}$ seem as real as possible to the discriminator. Unfortunately, after training, $f_\theta(\mathbf{x}, \mathbf{z})$ produces the same output for all values of $\mathbf{z}$ because of mode collapse, making conditional GANs ill-suited to the multimodal synthesis problem. Intuitively, this happens because making $\widehat{\mathbf{y}}$ as real as possible would push it towards the observed output $\mathbf{y}$, so the generator tries to make its output similar to the observed output $\mathbf{y}$ for all values of $\mathbf{z}$.

In [47], an alternative technique is proposed to train the generator network $f_\theta$, which is known as conditional IMLE (cIMLE). Rather than trying to make *all* outputs generated from different values of $\mathbf{z}$ similar to the observed output $\mathbf{y}$, it only tries to make *some* of them similar to the observed output $\mathbf{y}$. The generator is therefore only encouraged to map one value of $\mathbf{z}$ to the observed output $\mathbf{y}$, and reserve other values of $\mathbf{z}$ to *other* reasonable outputs that are not in the training dataset. This makes it possible to perform *multimodal* synthesis. Also, unlike cGANs, cIMLE does not use a discriminator and therefore does not require adversarial training, which makes training more stable. The following training objective takes the following form:

$$\min_\theta \mathbb{E}_{\mathbf{z}_{1,1},\ldots,\mathbf{z}_{n,m} \sim \mathcal{N}(0,\mathbf{I})} \left[ \sum_{i=1}^n \min_{j \in \{1,\ldots,m\}} d(f_\theta(\mathbf{x}_i, \mathbf{z}_{i,j}), \mathbf{y}_i) \right],$$

where $d(\cdot, \cdot)$ is a distance metric, $m$ is a hyperparameter, and $\mathbf{x}_i$ and $\mathbf{y}_i$ are the $i$th input and observed output in the dataset.

Unfortunately, in [47], a different generator network was used for each task, thereby limiting its applicability more generally to other tasks. Moreover, the generated outputs have low fidelity and lack fine details, especially when compared to unimodal methods like cGANs, raising the question of whether cIMLE can produce images of comparable fidelity to cGANs despite having no discriminator. In this paper, we address these issues and answer the latter question in the affirmative.

## 4 Method

### 4.1 Leveraging Multiple Scales

Images contain structure at different scales, and it is important to both leverage cues at multiple scales in the input image and produce realistic global structure and fine details in the output image. To this end, we devise a meta-architecture suited to modelling structure at multiple scales. This high-level theme of multi-scale processing isn't new and has been used by many methods in various contexts, e.g.: [15, 53, 35, 9, 56]. What is interesting is the precise way this is done in order to enable intermediate supervision and hierarchical sampling, which are described below. It turns out both are critically important to achieving high fidelity generation, which is validated by an ablation study in the appendix.

In our architecture (shown in Figure 1), we have a sequence of modules, each of which handles an input image of a particular resolution and outputs an image of the same resolution. We downsample the input image repeatedly by a factor of 2 to obtain a set of input images at different resolutions and feed them into different modules. Each module takes a latent code whose spatial dimensions correspond to its resolution and the upsampled output of the module for the next lowest resolution as input. Note that this architecture generalizes to varying levels of resolution, since we can simply add more modules for high-resolution outputs.

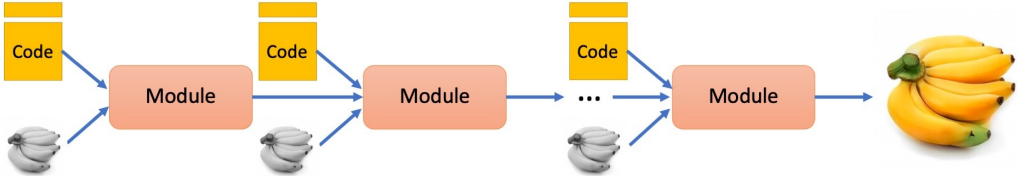

Figure 1: Our CAM-Net model consists of multiple modules, each of which operates on $2\times$ the resolution of previous one.

We add supervision to the output of each intermediate module to encourage similarity between the generated image and the real image. Effectively, the distance metric in cIMLE is chosen to be the *sum* over perceptual distances between the output of each module and the real image downsampled to the same resolution, rather than just the perceptual distance between the final output and the real image. We choose LPIPS [92] as our perceptual distance metric.

### 4.2 Hierarchical Sampling

Recall that for each input, cIMLE generates many samples and tries to make one of them similar to the observed output. The samples that are not selected correspond to the other possible outputs that are unobserved. So, the more samples that are generated during training, the more modes of the output distribution cIMLE can model. While we would ideally like to use many samples during training, generating samples is expensive, and so in practice, we can only generate just enough samples for cIMLE to learn effectively. This forces a tradeoff between the number of samples and performance, which is less than ideal.

To get around this conundrum, we propose a novel sampling strategy, known as hierarchical sampling. Because cIMLE only uses the sample that is closest to the observed output for training, the key idea is to sample close to the region of the latent code space that is likely to be close to the observed output. This avoids generating samples that are unlikely to be selected, thereby increasing the effective number of samples without actually generating all of them.

To this end, we generate samples for different modules successively, each of which operates at different resolutions. In the first stage, we sample latent codes for the first module, generate low-resolution images from them using the first module and select the latent code whose generated image is closest to the observed output downsampled to the appropriate resolution. In the next stage, we condition on the latent code of first module by setting it to the latent code selected in the previous stage. We sample latent codes for the second module and generate images at the next higher resolution from them using the first and second modules. In subsequent stages, we repeat the analogous procedure for the later modules. Note that this procedure is only used at train time; at test time, the latent codes for different modules are sampled independently because the goal at test time is to generate all possible outputs, including those that are unobserved.

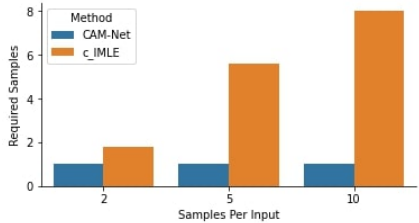

Figure 2: Comparison of sample efficiency of hierarchical sampling (HS) to vanilla sampling (which samples latent codes for different modules independently). The relative disparity of the required number of samples needed to achieve the same LPIPS distance to the observed output with/without HS is shown, where the number of required samples for HS is normalized to 1. The reported results are averaged over 10 independent runs. As shown, as the number of samples used per module increases in the case of HS, more samples are needed by vanilla sampling to match the distance attained by HS.

We validate the improved sample efficiency of hierarchical sampling in Figure 2. We compare the number of samples required to obtain the same level of LPIPS distance to the observed output, with and without hierarchical sampling. As shown, vanilla sampling requires 2 to 8 times more samples than hierarchical sampling to reach the same LPIPS distance, and the difference becomes larger as the number of samples used for each module in hierarchical sampling increases.

### 4.3 Unified Model Architecture

In the generative modelling literature, architecture design has played an important role in advancing image fidelity [58, 60, 75], and different types of generative models have different optimal choices of architecture due to differences in the goal (mode seeking vs. mode covering) and training objective (adversarial vs. non-adversarial).

Prior cIMLE architectures are unable to generate fine details, so to generate high fidelity images, we design a new architecture for cIMLE. Unlike prior cIMLE architectures [47], the proposed architecture performs well across a broad variety of image synthesis tasks; in fact, the single proposed architecture significantly outperforms prior task-specific architectures, as shown later in Sect. 6.

In each module, the backbone architecture comprises of two branches, a main branch consisting of residual-in-residual dense blocks (RRDB) [79] and an auxiliary branch consisting of a sequence of dense layers, known as a mapping network [38], that produces scaling factors and offsets for different channels in output produced by each RRDB. In Figure 4, we show the inner workings of each RRDB, which is made up of dense blocks and residual connections. Unlike traditional RRDB, which comes without normalization and deliberately removes batch normalization in particular, we apply weight normalization [64] to all convolution layers.

The precise design of the architecture is essential to achieving high image fidelity across various tasks. For example, we found the default number of blocks and channels in [79] to work poorly with IMLE, and needed to increase the number of channels and decrease the number of blocks at the same time. Arriving at the sweet spot in the space of architectures required thorough experimentation. As we will show in Sect. 6 and the appendix, the combination of the design motifs leads to a substantial improvement in generated image fidelity compared to prior architectures. We also include an ablation study of different components to show the effectiveness of each in the appendix.

## 5 Tasks

We apply our single proposed model to four very different conditional image synthesis tasks, namely single image super-resolution, image colourization, image synthesis from scene layouts and image decompression. We discuss the details of each task below.

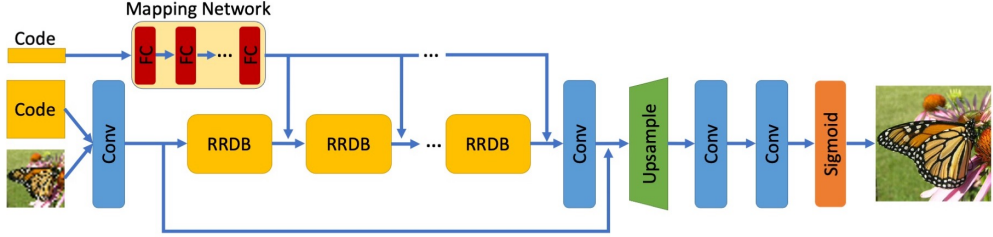

Figure 3: Details of the architecture backbone. See Figure 4a for the inner workings of RRDB blocks.

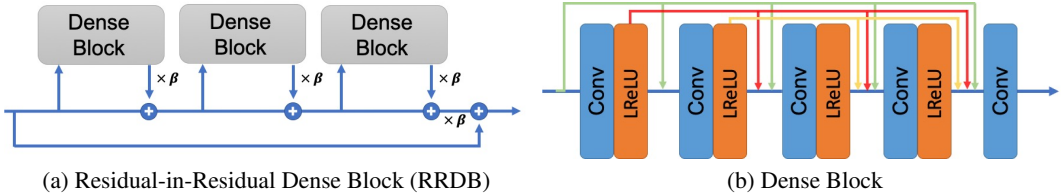

(a) Residual-in-Residual Dense Block (RRDB)  (b) Dense Block

Figure 4: (a) Inner workings of Residual-in-Residual Dense Blocks (RRDBs), which comprises of dense blocks (details in (b)). $\beta$ is the residual scaling parameter. (b) Inner workings of dense blocks.

## 5.1 $16\times$ Super-Resolution

Given a low-resolution input image, the goal is to generate different possible higher-resolution output images that are all geometrically consistent with the input. Applications include photo enhancement, remote sensing and medical imaging. Unlike most methods that consider relatively small upscaling factors of $2-4\times$, we apply our method to a much more challenging [6] setting of $16\times$ upscaling. Our multimodal method is a good fit for this setting since there could be multiple high-resolution images with perceptible differences in details that correspond to the same low-resolution image.

Super-resolution at such a high upscaling factor requires learning the semantics of the objects depicted in images. Hence, we use ILSVRC-2012 dataset which categorizes the images by semantics. Three categories that consist of 3900 images are selected and anisotropically downsampled to $512\times512$ which are used as target output images. We further downsample them with a bicubic kernel to $32\times32$ to arrive at the input images. We pick the leading IMLE-based method [47], labelled as cIMLE, and the leading $16\times$ GAN-based method that is dedicated to super-resolution, RFB-ESRGAN [66] as baselines for this task.

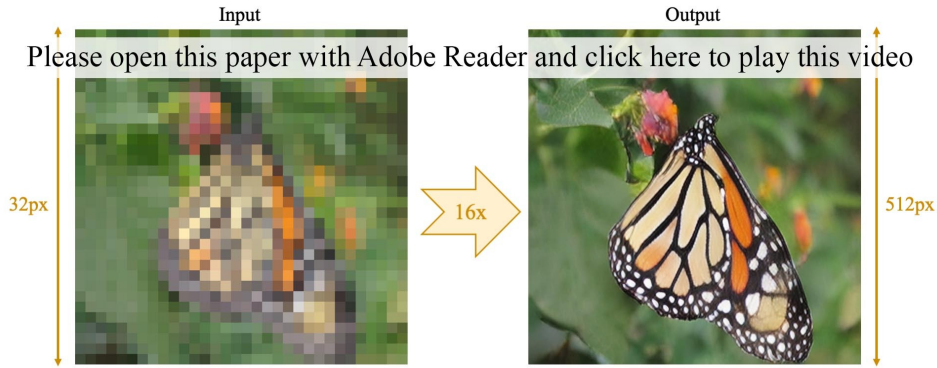

Figure 5: Visualization of different samples generated by our method (CAM-Net) and the input for super-resolution. As shown, CAM-Net generates different high quality textures for example on the edge of the butterfly's wing.

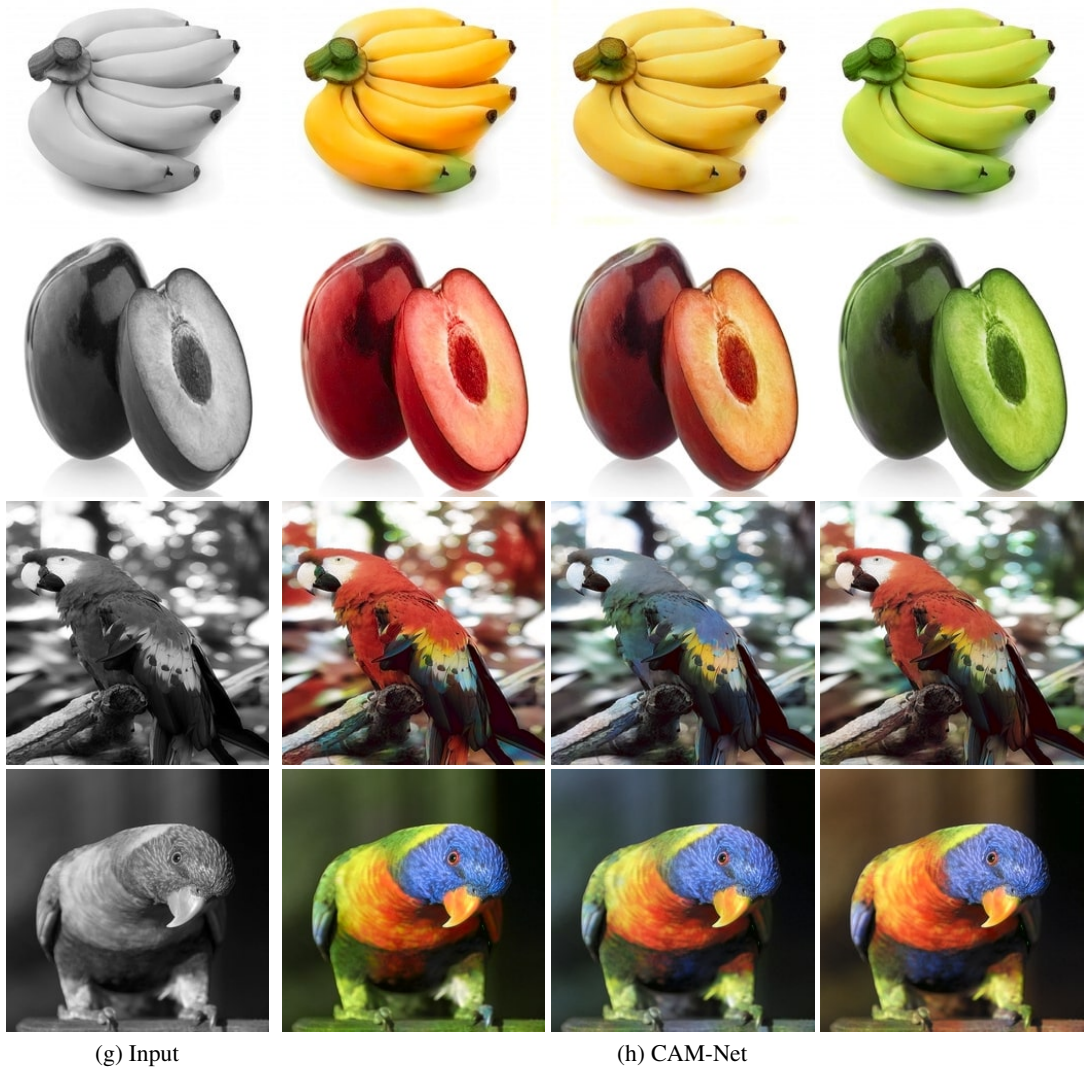

(g) Input                           (h) CAM-Net

Figure 6: Visualization of different samples generated by our method (CAM-Net) and the input for image colourization. As shown in the figure, in addition to common colours, CAM-Net also produces a variety of colours, such as green bananas and plums. Similarly, generating parrots with different body colours also shows the power of CAM-Net in terms of multimodality.

## 5.2 Image colourization

Given a grayscale image, the goal is to generate colours for each pixel that are consistent with the content of the input image. Applications include the restoration of old photos, for example those captured by black-and-white cameras or those captured by colour cameras that has become washed out over time. Since there could be many plausible colourings of the same grayscale image, the goal of multimodal synthesis is to generate different plausible colourings, which would provide the user the ability to choose the preferred colouring among them.

We compare our method to cIMLE and leading methods specifically designed for colourization, such as Zhang et al. [91], Iizuka et al. [30] and Larsson et al. [44] on the Natural Colour Dataset (NCD) [4] and two categories from ILSVRC-2012.

## 5.3 Image Synthesis From Scene Layouts

Given a semantic segmentation map, the goal is to synthesize realistic images from the segmentation map. This task is challenging as the category label and the shape of each segment are the only cues to the model; in particular, no information about the appearance is provided to the model. There are many scenes with the same layout that correspond to the same segmentation map. So, multimodal synthesis aims to generate multiple images with the same scene layout and different appearances.

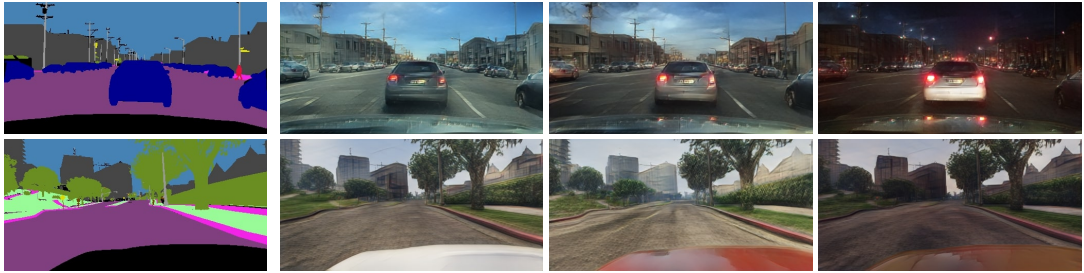

(c) Input                                    (d) CAM-Net

Figure 7: Visualization of different samples generated by our method (CAM-Net) and the input for image synthesis from semantic segmentation map. The first row is generated by the model trained on BDD100K, and the second row is generated by the model trained on GTA-5. As shown, CAM-Net is able to produce high-quality and highly detailed results at different times of a day; in particular, the rear of the car, the hood, the colour of the sky, and the lines of the street are examples of this.

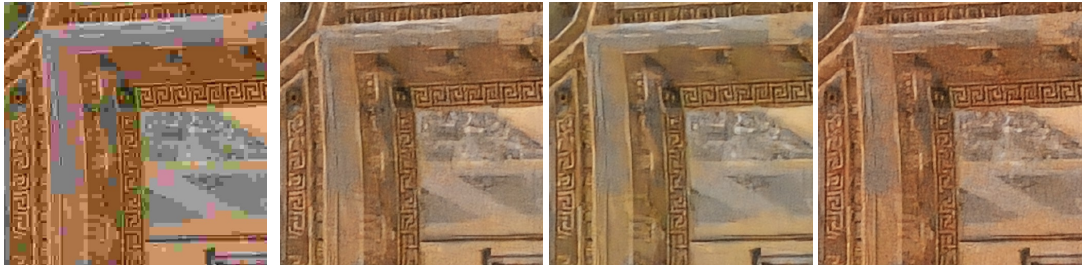

(a) Input                                    (b) CAM-Net

Figure 8: Visualization of different samples generated by our method (CAM-Net) and the input for image decompression. As shown, CAM-Net output successfully removes most artifacts and predicts diverse textures.

We evaluate on the BDD100K [86] and GTA-5 [61] datasets. Following prior work [47], we apply dataset and loss rebalancing to compensate for the data imbalance of the datasets.

### 5.4 Image Decompression

Given an image compressed aggressively with a standard lossy codec (e.g.: JPEG), the goal is to recover the original uncompressed image. Note that this task is different from image compression, where both the encoding and decoding model can be learned. Here, the encoding model is fixed (e.g.: JPEG), and only the decoding model is learned.

Image decompression is of practical interest since most images are saved in lossy compressed formats; noticeable artifacts may have been introduced during compression and the original uncompressed images have been lost. It would be nice to have the capability to recover an image free of artifacts. Because compression causes irreversible information loss, multiple artifact-free images are possible. With user guidance, the most preferred version can be selected and saved for future use.

We compare our method to DnCNN [88] and cIMLE on RAISE1K [14] dataset. We consider a challenging setting with substantial information loss, where the compressed images are compressed aggressively with JPEG at 1% quality.

#### 5.4.1 Training Details

We use a four-module design for all tasks; except for image synthesis from scene layouts, we adopt a five-module design. Input to each module is downsampled anisotropically from the full resolution input to fit the corresponding resolution, except for super-resolution, only the first module take in the low-resolution input. All models were trained using a NVIDIA V100 GPU on an internal cluster.

|  | Super-Resolution | | | Image Decompression | | |
| --- | --- | --- | --- | --- | --- | --- |
|  | *CAM-Net* | *cIMLE* | *RFB-ESRGAN* | *CAM-Net* | *cIMLE* | *DnCNN* |
| FID | **16.75** | 27.34 | 19.56 | **72.75** | 100.48 | 109.38 |

|  | colourization | | | | Scene Layout Synthesis | |
| --- | --- | --- | --- | --- | --- | --- |
|  | *CAM-Net* | *cIMLE* | *Zhang et al.* | *Iizuka et al.* | *Larsson et al.* | *CAM-Net* | *cIMLE* |
| FID | **33.19** | 36.38 | 57.95 | 85.88 | 47.44 | **48.43** | 88.53 |

Table 1: Comparison of Fréchet Inception Distance (FID) to the target of the samples generated by our method (CAM-Net) and the baselines on different tasks. Lower values of FID are better. We compare favourably on this perceptual metric (FID).

|  | Super-Resolution | | Image Decompression | |
| --- | --- | --- | --- | --- |
| $\sigma$ | *CAM-Net* | *cIMLE* | *CAM-Net* | *cIMLE* |
| 0.3 | $\mathbf{5.72 \times 10^{-2}}$ | $5.48 \times 10^{-2}$ | $\mathbf{5.13 \times 10^{-2}}$ | $4.93 \times 10^{-2}$ |
| 0.2 | $\mathbf{5.86 \times 10^{-3}}$ | $5.22 \times 10^{-3}$ | $\mathbf{3.80 \times 10^{-3}}$ | $3.14 \times 10^{-3}$ |
| 0.15 | $\mathbf{3.44 \times 10^{-4}}$ | $2.73 \times 10^{-4}$ | $\mathbf{2.23 \times 10^{-4}}$ | $1.32 \times 10^{-4}$ |

|  | colourization | | Scene Layout Synthesis | |
| --- | --- | --- | --- | --- |
| $\sigma$ | *CAM-Net* | *cIMLE* | *CAM-Net* | *cIMLE* |
| 0.3 | $\mathbf{1.24 \times 10^{-1}}$ | $1.05 \times 10^{-1}$ | $\mathbf{2.58 \times 10^{-2}}$ | $1.71 \times 10^{-2}$ |
| 0.2 | $\mathbf{6.21 \times 10^{-2}}$ | $4.56 \times 10^{-2}$ | $\mathbf{1.00 \times 10^{-3}}$ | $4.91 \times 10^{-4}$ |
| 0.15 | $\mathbf{2.84 \times 10^{-2}}$ | $1.79 \times 10^{-2}$ | $\mathbf{1.45 \times 10^{-4}}$ | $4.5 \times 10^{-5}$ |

Table 2: Comparison of faithfulness weighted variance of the samples generated by our method (CAM-Net) and conditional IMLE (cIMLE) on different tasks. Higher value shows more variation in the generated samples that are faithful to the original image. $\sigma$ is the bandwidth parameter for the Gaussian kernel used to compute the faithfulness weights.

## 6 Results

### 6.1 Quantitative Results

We evaluate our results along two axes, perceptual quality and output diversity. Perceptual quality is measured using the Fréchet Inception Distance (FID) [25]; classical metrics like PSNR and SSIM do not capture perceptual quality well [45] and therefore are ill-suited to this setting. Output diversity is measured using faithfulness-weighted variance [47], which is the LPIPS distance between the output samples conditioned on the same input and the mean of the samples, weighted by the consistency with the target output measured by a Gaussian kernel. The kernel bandwidth parameter $\sigma$ trades off the importance of consistency vs. diversity.

We compare the perceptual quality and output diversity in Tables 1 and 2. As shown in Table 1, CAM-Net outperforms both the multimodal baseline, cIMLE, and specialized unimodal baselines, in terms of FID. As shown in Table 2, CAM-Net outperforms the multimodal baseline, cIMLE, in terms of faithfulness-weighted variance at all bandwidth parameters for all tasks. Comparisons to unimodal baselines are omitted because their faithfulness-weighted variances are zero. These comparisons indicate that CAM-Net is able to produce more realistic and diverse images than the baselines.

### 6.2 Qualitative Results

We show the results of our method and the input for super-resolution in Figure 5, colourization in Figure 6, image synthesis from scene layouts in Figure 7, image decompression in Figure 8 and comparisons to the baselines in the appendix. As shown, CAM-Net generates high quality and diverse results.

# 7 Limitation

Our method can only generate diversity that is present in the training data. Take colourization as an example, if no green apples are observed anywhere in the training data, then our model cannot produce green apples.

# 8 Conclusion

In this paper, we developed an improved method for the challenging problem of multimodal conditional image synthesis based on the conditional IMLE (cIMLE) framework. We addressed three main issues of the prior cIMLE-based approach: image fidelity, task-specific architectures and sample efficiency. We proposed a new method, CAM-Net, which is a single architecture that can be applied to a broad variety of tasks. The modularity of the architecture allows us to devise a novel hierarchical sampling scheme for IMLE, which improves the sample efficiency. Finally, we demonstrate our single architecture can generate significantly higher fidelity output images without compromising on diversity.

# 9 Societal Impact

Because CAM-Net can be applied to a broad range of conditional image synthesis tasks, its societal impact depends on the applications that it is used for. While most applications are harmless, it could be potentially used for tasks like demosaicing which have privacy or copyright implications. In addition, since CAM-Net adopts a modular design, more modules could be added by downstream users to obtain more refined results for particular tasks, which can result in a larger model with more parameters. Such a model may require more compute, which can increase carbon dioxide emissions.

[94] Jun-Yan Zhu, Taesung Park, Phillip Isola, and Alexei A. Efros. Unpaired image-to-image translation using cycle-consistent adversarial networks. *2017 IEEE International Conference on Computer Vision (ICCV)*, pages 2242–2251, 2017.

[95] Jun-Yan Zhu, Richard Zhang, Deepak Pathak, Trevor Darrell, Alexei A. Efros, O. Wang, and E. Shechtman. Toward multimodal image-to-image translation. In *NIPS*, 2017.

[96] Jun-Yan Zhu, Richard Zhang, Deepak Pathak, Trevor Darrell, Alexei A Efros, Oliver Wang, and Eli Shechtman. Toward multimodal image-to-image translation. In *Advances in Neural Information Processing Systems*, pages 465–476, 2017.
15

## A  More Results

We include more results on $16\times$ super-resolution, colourization, image synthesis from scene layouts and image decompression in the following pages. Videos of these results are also available as ancillary files on arXiv.

### A.1  $16\times$ **Super-Resolution Results**

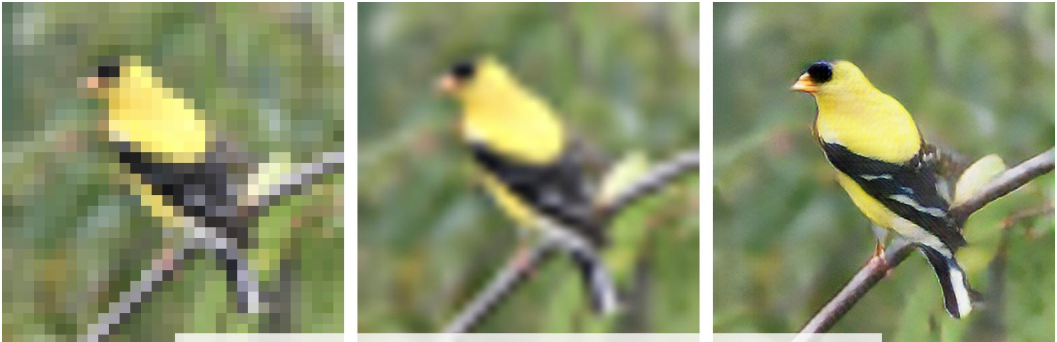

(a) Input  (b) Bicubic Upsampling  (c) ESRGAN

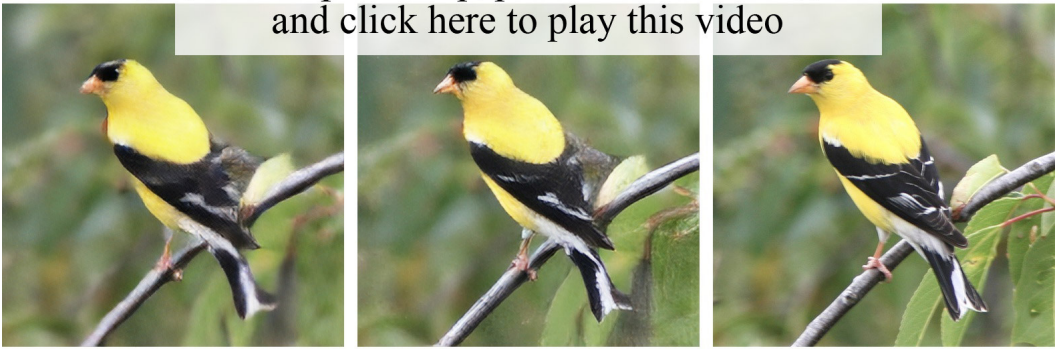

(d) cIMLE  (e) CAM-Net  (f) Original Image

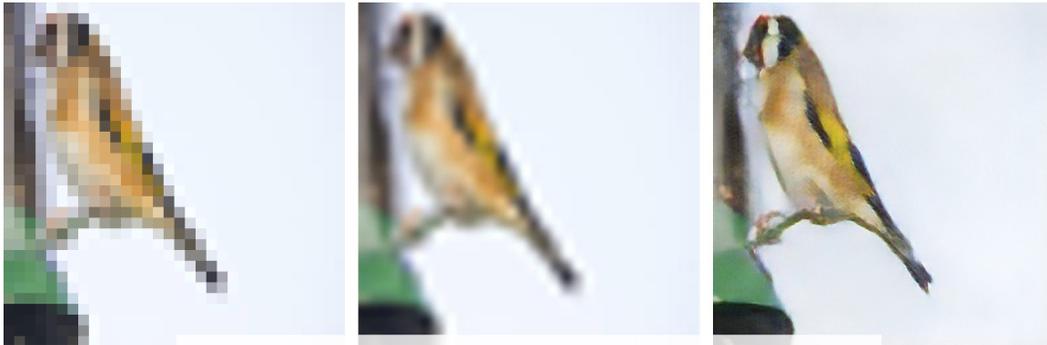
(a) Input        (b) Bicubic        (c) ESRGAN

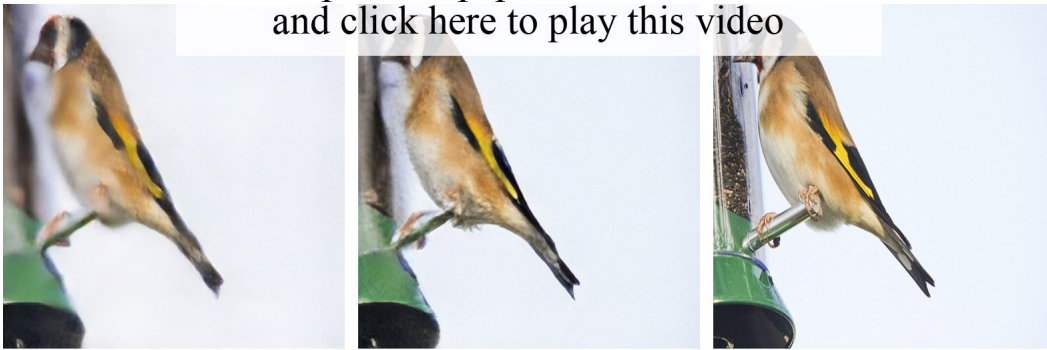
(d) cIMLE        (e) CAM-Net        (f) Original Image

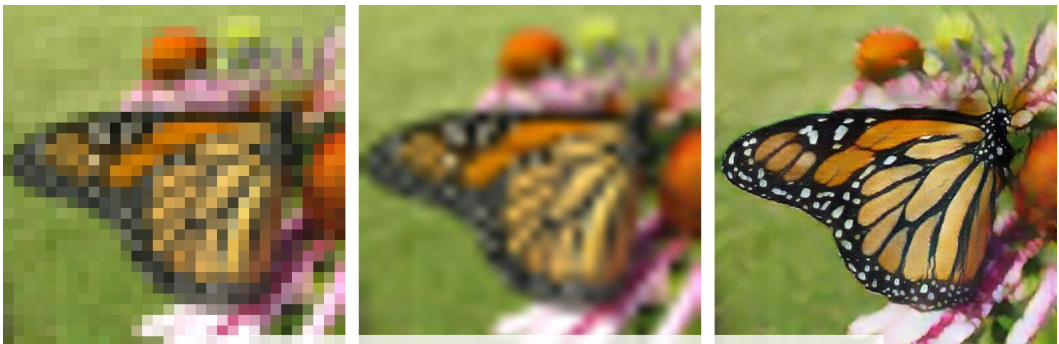
(a) Input        (b) Bicubic        (c) ESRGAN

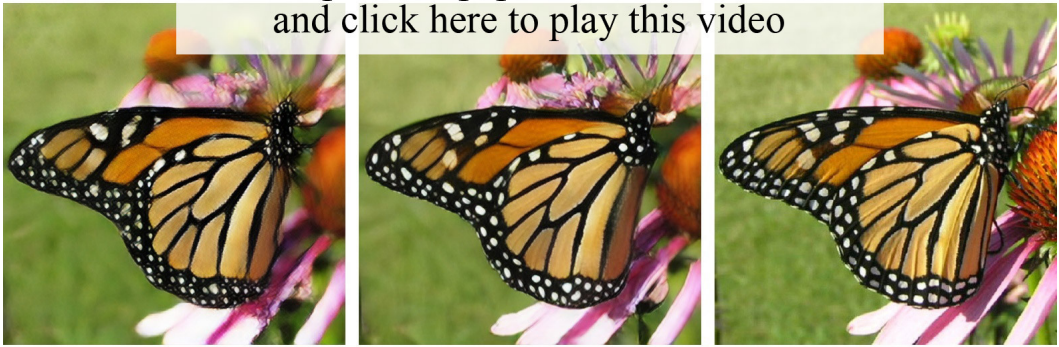
(d) cIMLE        (e) CAM-Net        (f) Original Image

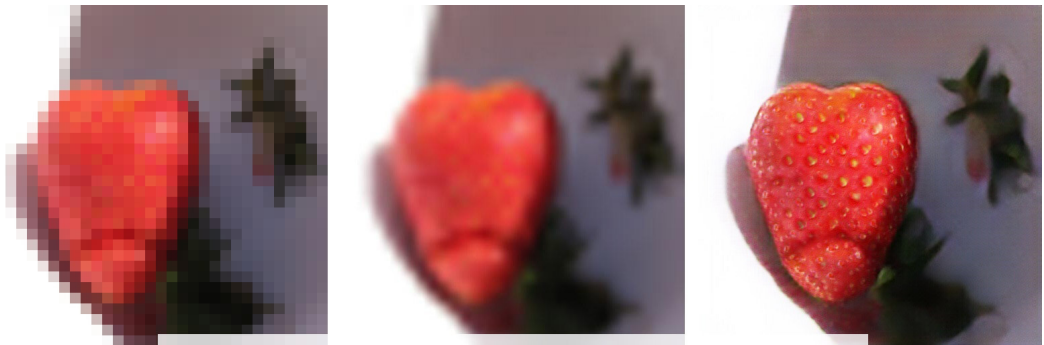
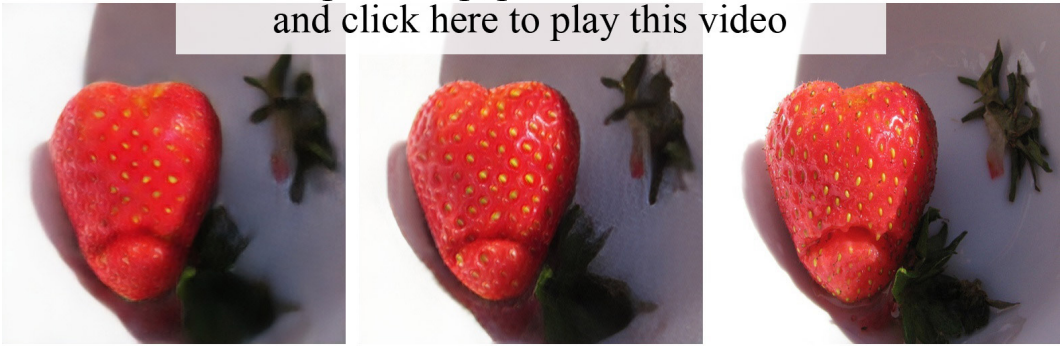

(a) Input          ESRGAN

(d) cIMLE          (e) CAM-Net          (f) Original Image

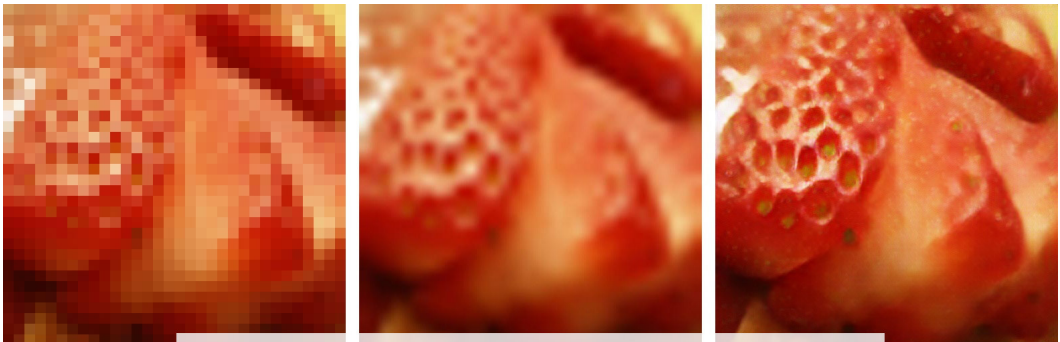
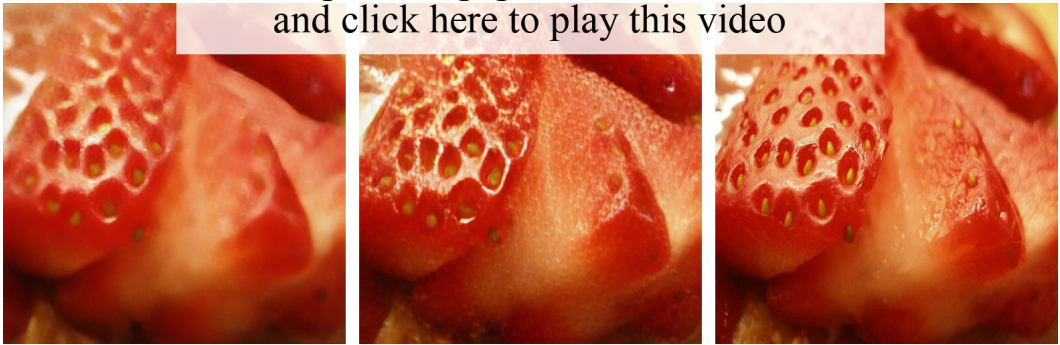

(a) Input          ESRGAN

(d) cIMLE          (e) CAM-Net          (f) Original Image

## A.2 Image colourization Results

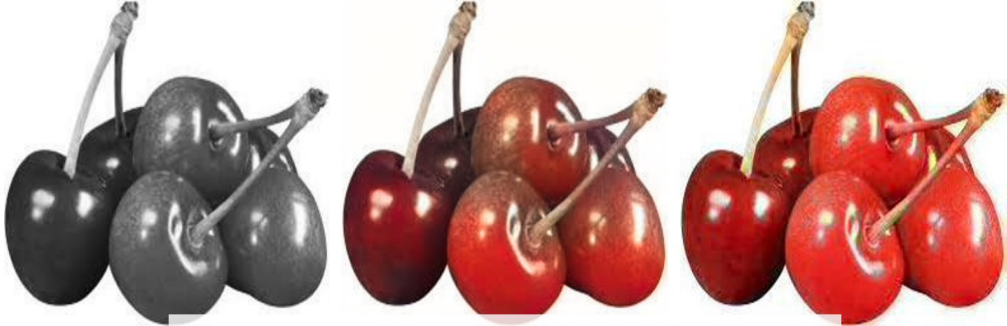

(a) Input  (b)  (c) Zhang et al.

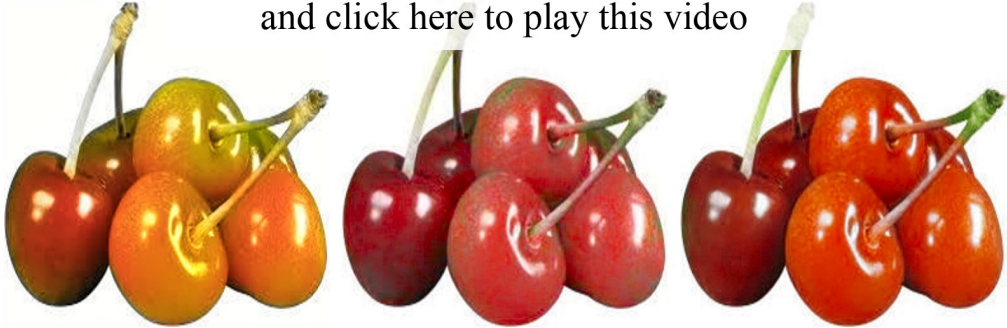

(d) Larsson et al.　　　(e) cIMLE　　　(f) CAM-Net

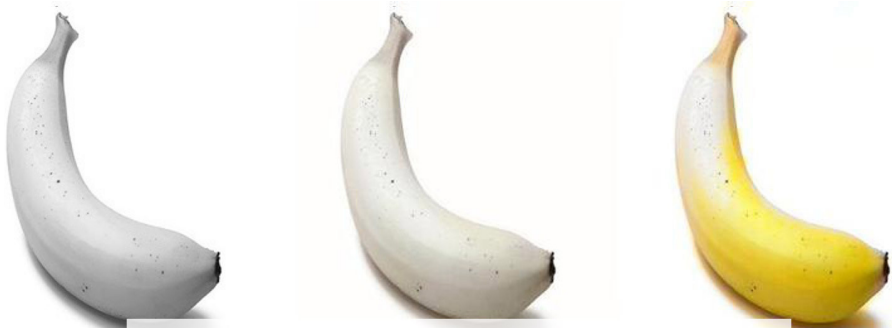

(a) Input  (b)  (c) Zhang et al.

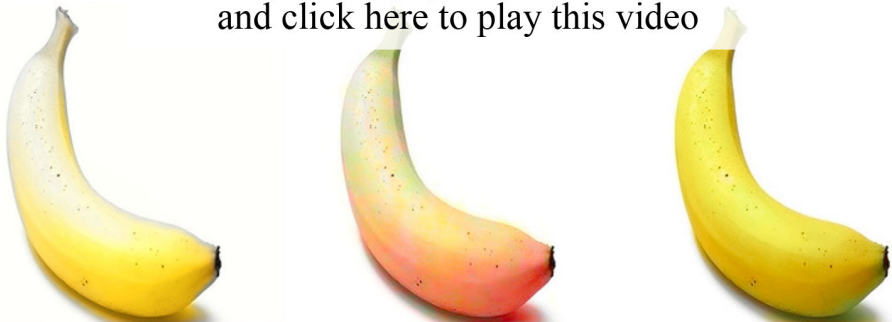

(d) Larsson et al.　　　(e) cIMLE　　　(f) CAM-Net

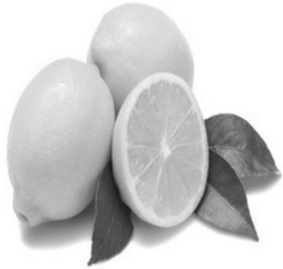
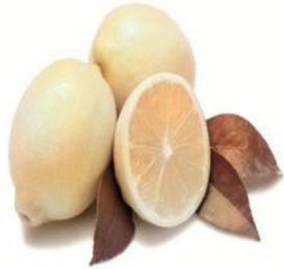
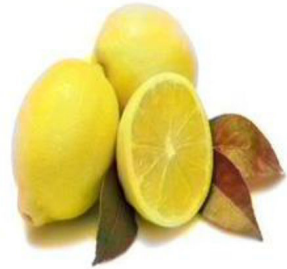

(a) Input      (b) Zhang et al.      (c) Zhang et al.

Please open this paper with Adobe Reader and click here to play this video

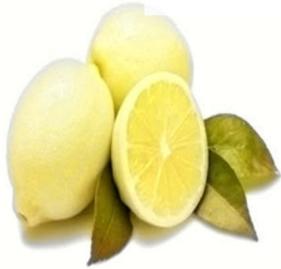
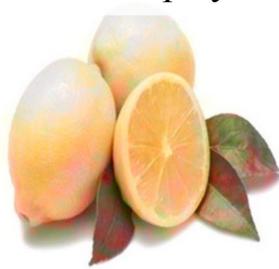
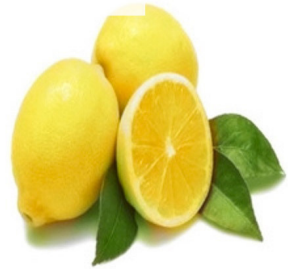

(d) Larsson et al.      (e) cIMLE      (f) CAM-Net

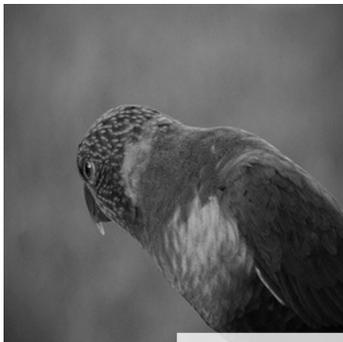
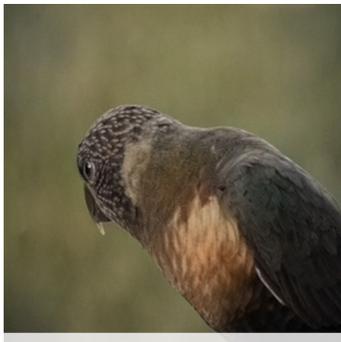
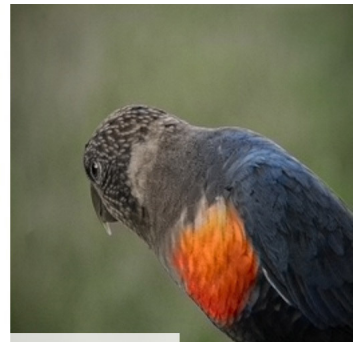

(a) Input      (b) Zhang et al.      (c) Zhang et al.

Please open this paper with Adobe Reader and click here to play this video

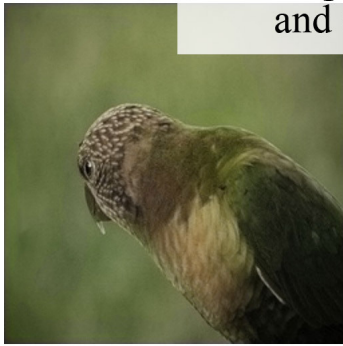
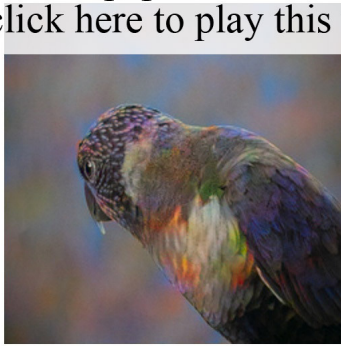
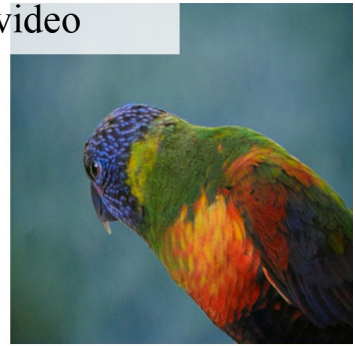

(d) Larsson et al.      (e) cIMLE      (f) CAM-Net

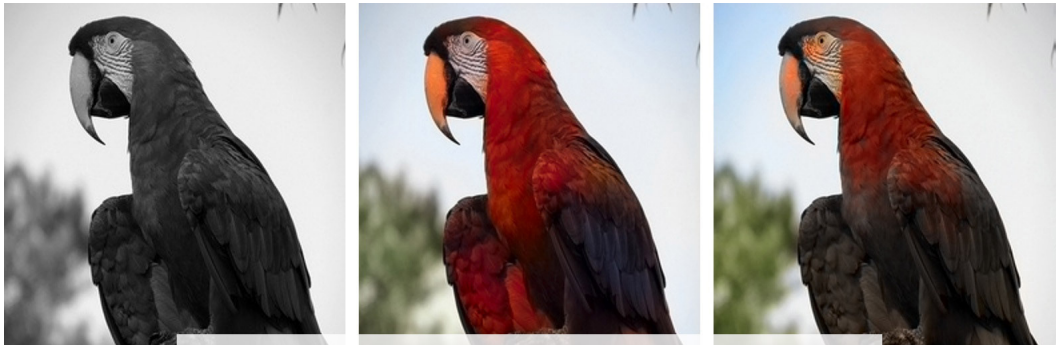
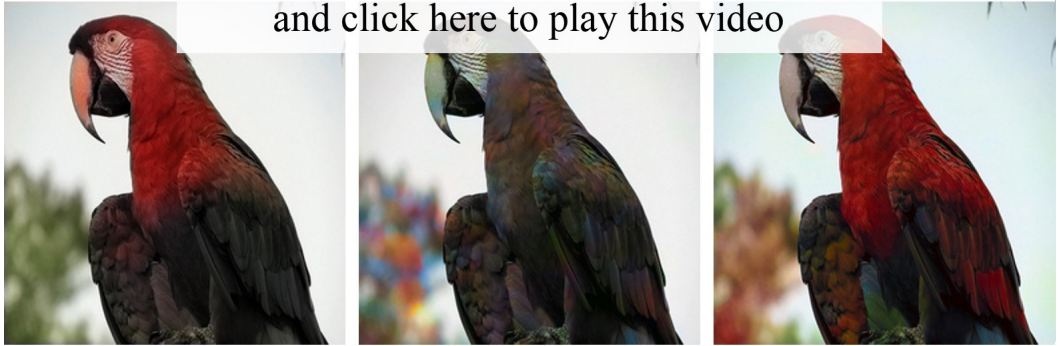

(a) Input (b) Zhang et al. (c) Zhang et al.
(d) Larsson et al. (e) cIMLE (f) CAM-Net

Please open this paper with Adobe Reader and click here to play this video

### A.2.1 Multimodality Comparison To Zhang et al.

In Zhang et al. [91], image colourization is formulated as a classification problem by discretizing the colour space and predicting a discrete distribution over the different bins. This models the *marginal* distribution over colours for each individual pixel, but not the *joint* distribution over combinations of colours for all pixels.

We can modify the method of Zhang et al. to produce multiple outputs by sampling from the predicted marginal distribution for each pixel independently. Below we visualize the results produced by this approach and compare to the results of our method.

As shown in Figure 9, Zhang et al. fails to produce spatially consistent colourizations - for example, some regions are coloured orange or yellow, while others are coloured yellow. While orange, yellow and green pears are all possible (indicating that the *marginal* distribution is modelled correctly), a pear cannot have orange, yellow and green spots at the same time. This occurs because Zhang et al. does not model the *joint* distribution over the colours of different pixels and so does not learn the correlations between the colours of different pixels. On the other hand, CAM-Net produces spatially consistent colourizations, because it learns the joint distributions.

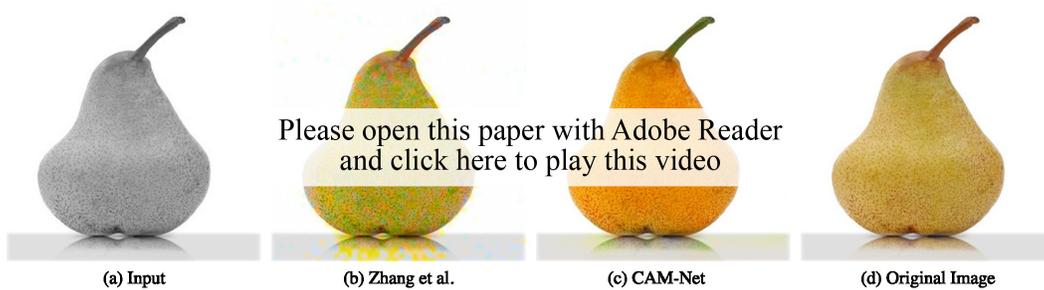

Figure 9: Click on the image to see diversity comparison between Zhang et al. and CAM-Net. As shown, CAM-Net produces more spatially consistent results than Zhang et al.

## A.3 Scene Synthesis Results

### A.3.1 BDD100K

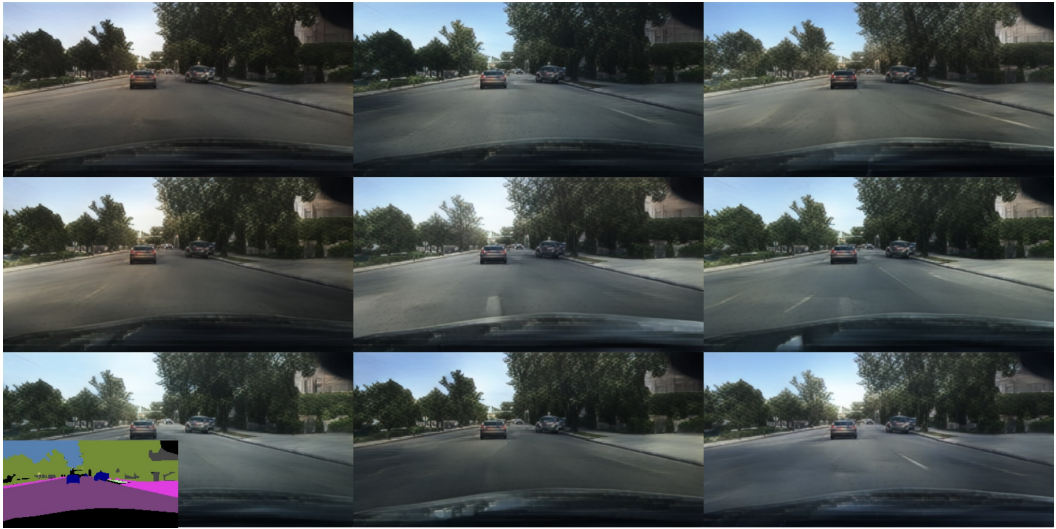

(a) CAM-Net

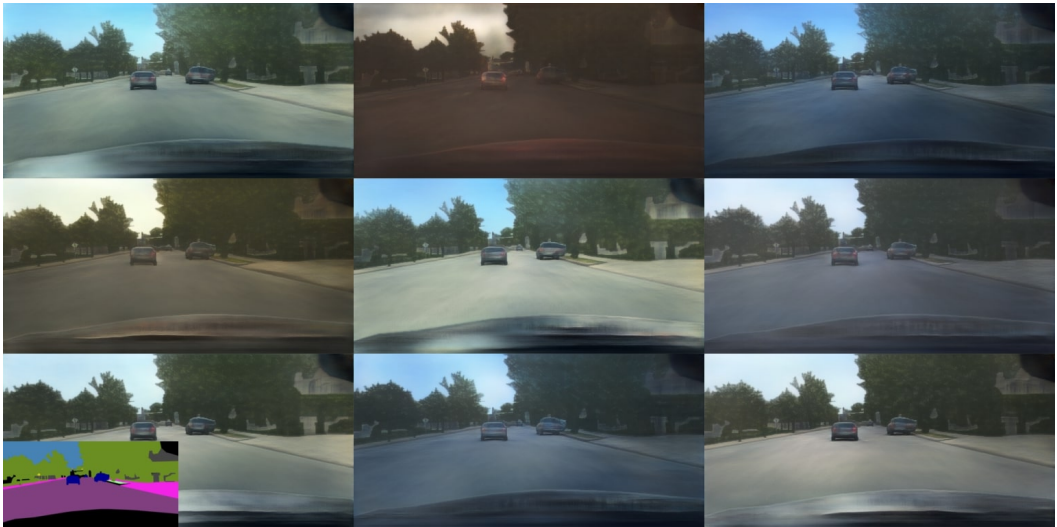

(b) cIMLE

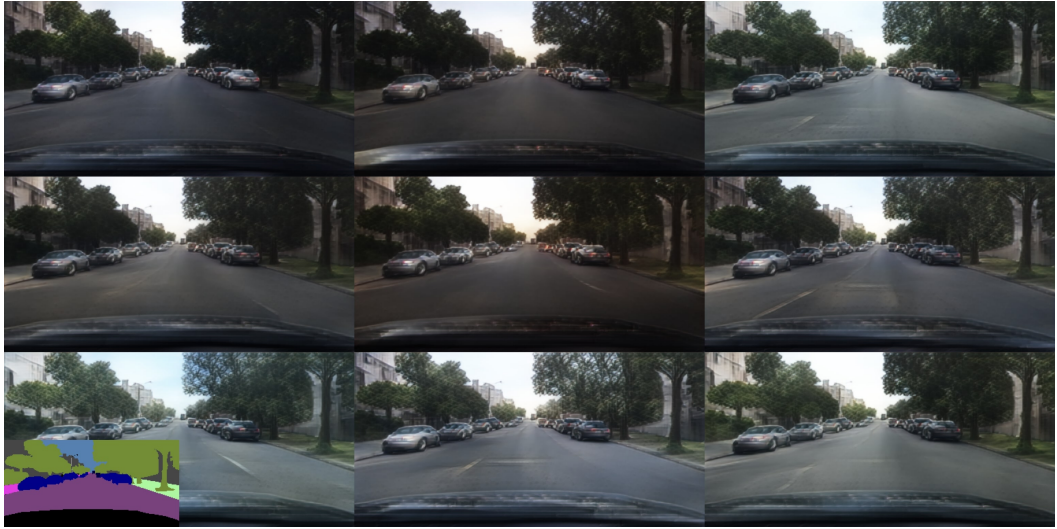

(a) CAM-Net

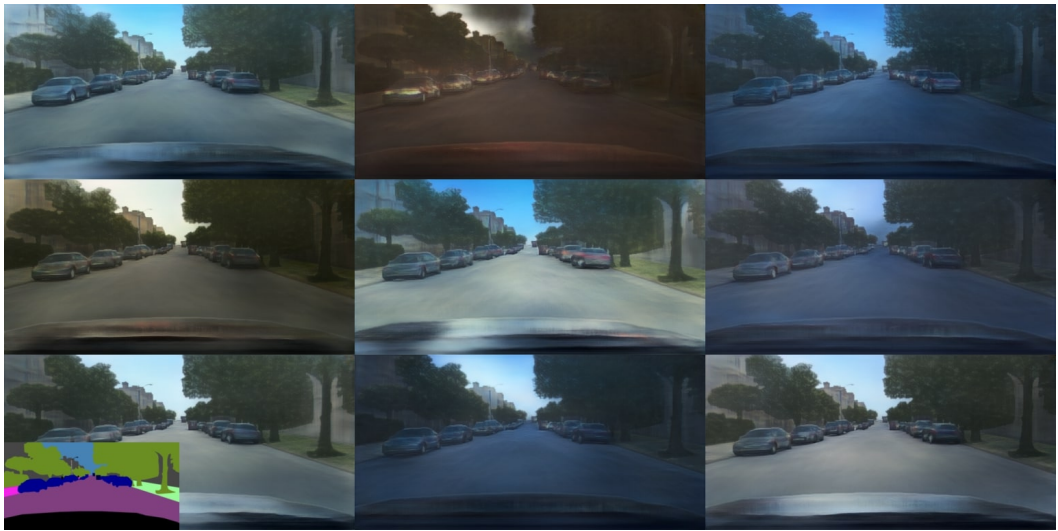

(b) cIMLE

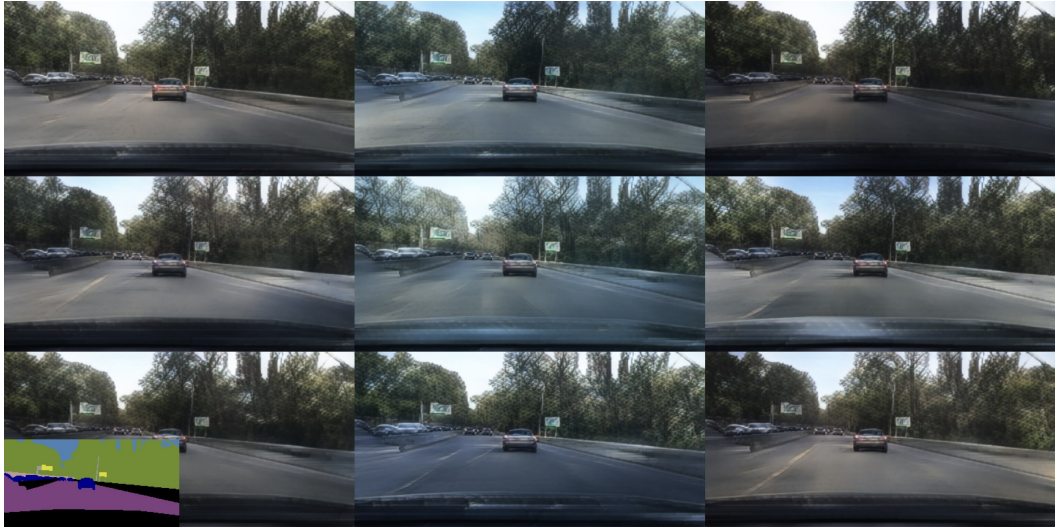

(a) CAM-Net

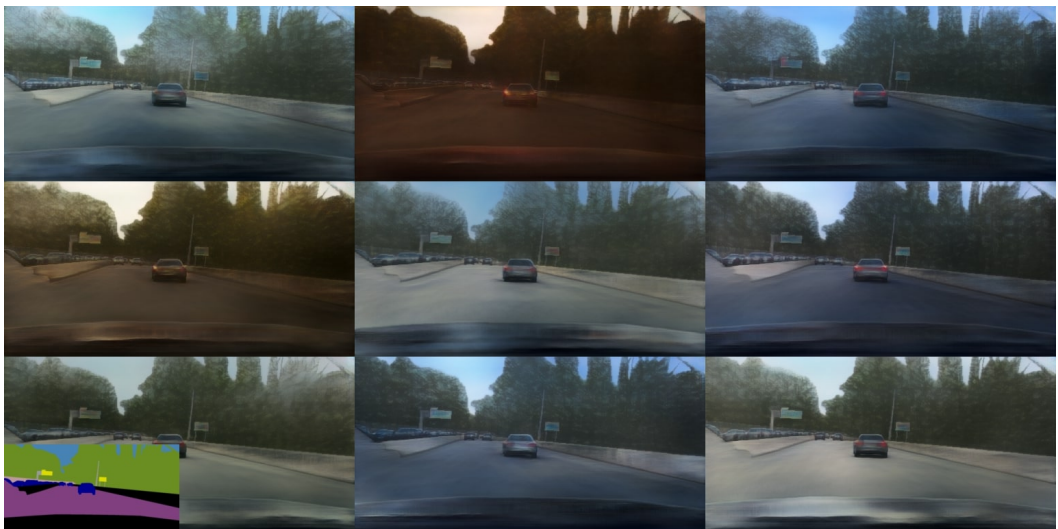

(b) cIMLE

### A.3.2 GTA-5

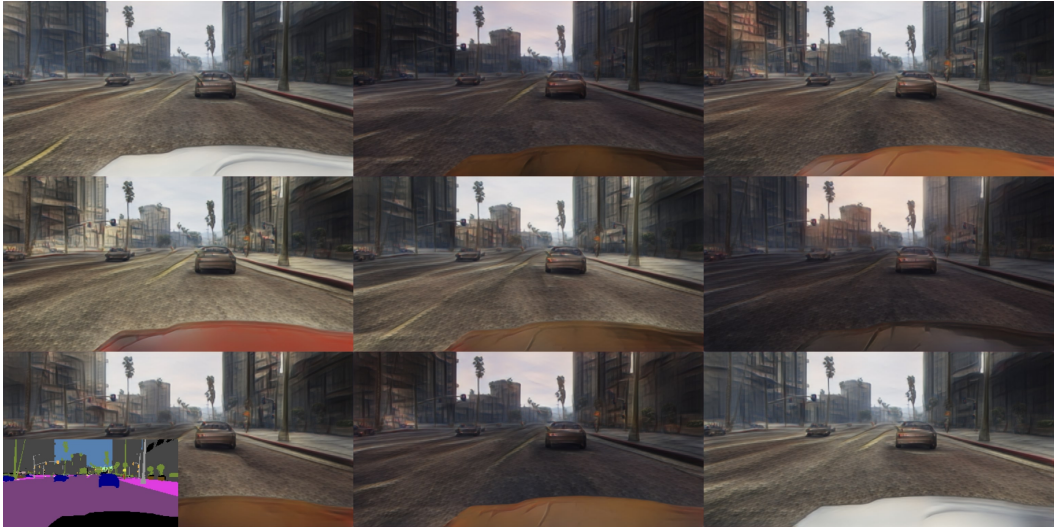

(a) CAM-Net

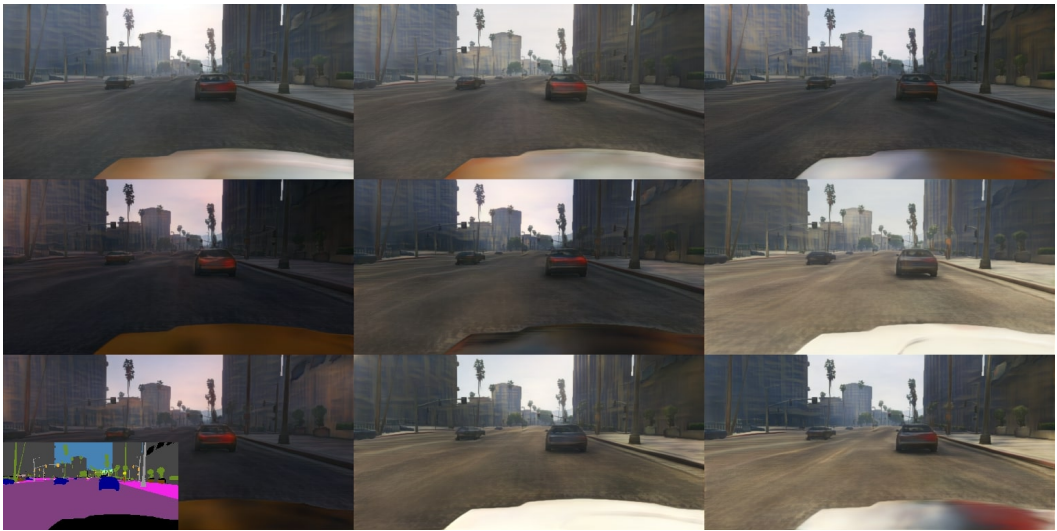

(b) cIMLE

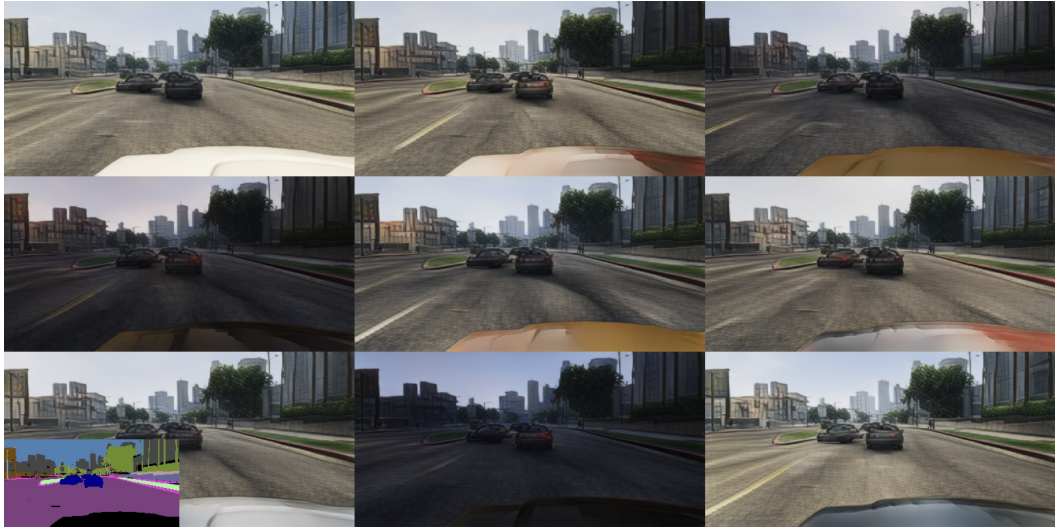

(a) CAM-Net

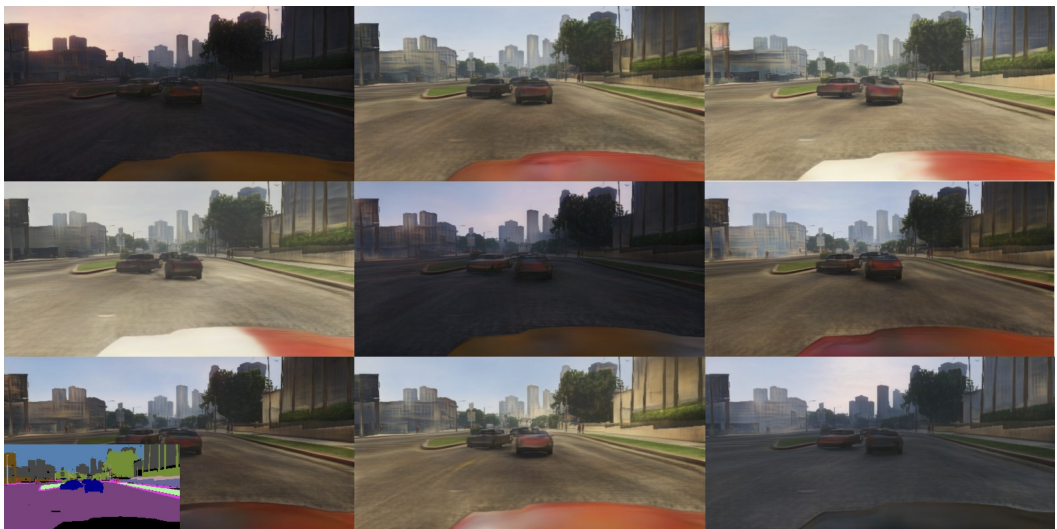

(b) cIMLE

## A.4 Image Decompression Results

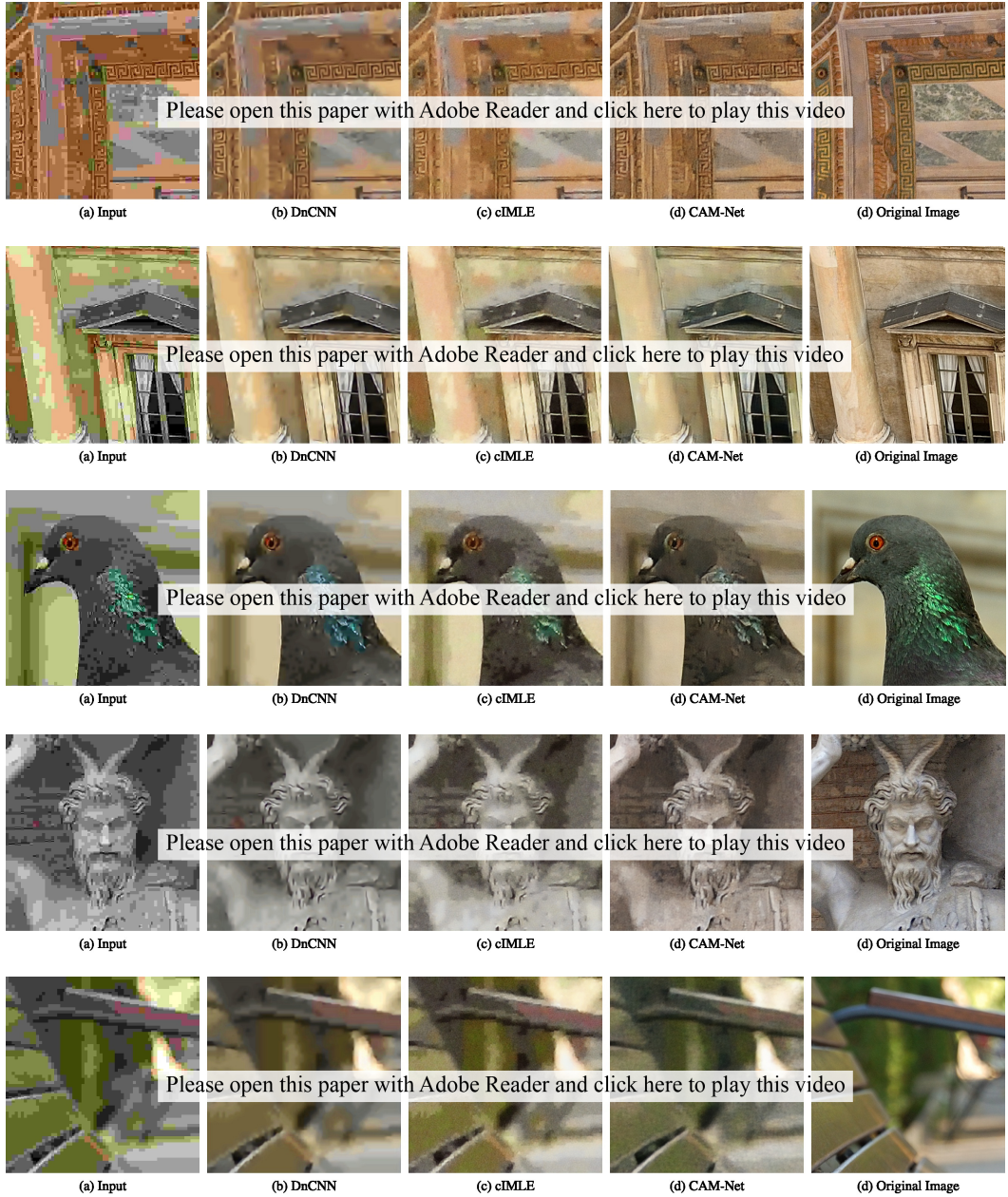

## B  Ablation Study

We incrementally remove (1) hierarchical sampling (HS), (2) mapping network (MN), (3) intermediate supervision (IS), (4) weight normalization (WN). As shown in Figure 10 and Table 3, each component is critical to achieving best results.

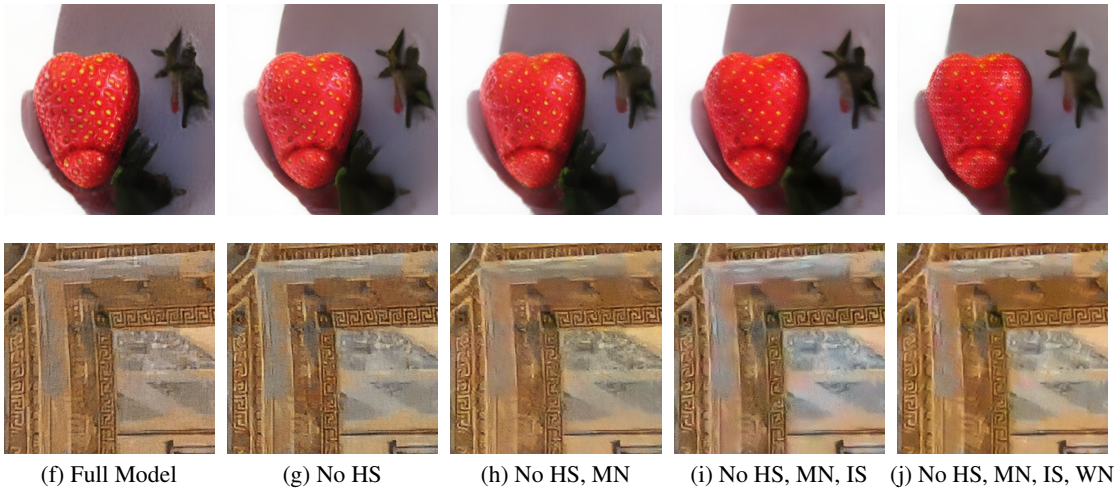

(f) Full Model    (g) No HS    (h) No HS, MN    (i) No HS, MN, IS    (j) No HS, MN, IS, WN

Figure 10: Visualization comparison of two tasks: super-resolution (first row) and image decompression (second row) as we gradually remove (1) hierarchical sampling (HS), (2) mapping network (MN), (3) intermediate supervision (IS), (4) weight normalization (WN).

|    | Full Model | No HS | No HS, MN | No HS, MN, IS | No HS, MN, IS, WN |
|----|-----------|-------|-----------|---------------|-------------------|
| SR | 16.75     | 17.70 | 18.58     | 21.66         | 22.51             |
| DC | 72.75     | 81.24 | 84.57     | 99.54         | 102.82            |

Table 3: Comparison of Fréchet Inception Distance (FID) of two tasks: super-resolution (SR) and image decompression (DC) by gradually remove (1) hierarchical sampling (HS), (2) mapping network (MN), (3) intermediate supervision (IS), (4) weight normalization (WN).

## C  Training Stability

In Figure 11, we visualize the output of CAM-Net for a test input image over the course of training. As shown, the output quality improves steadily during training, thereby demonstrating training stability.
29

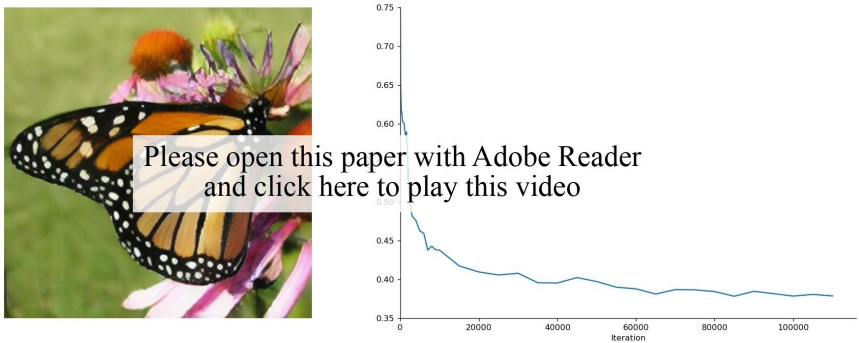

Figure 11: Click on the image to see output of model while it trains, demonstrating stable training. Video also available in supplementary materials.